\documentclass{article}
\usepackage{ijcai19}

\usepackage{times}
\usepackage{soul}
\usepackage{url}
\usepackage[hidelinks]{hyperref}
\usepackage[utf8]{inputenc}
\usepackage[small]{caption}
\usepackage{graphicx}
\usepackage{amsmath}
\usepackage{booktabs}
\usepackage{algorithm}
\usepackage{algorithmic}
\usepackage{natbib}
\usepackage{multirow}
\usepackage{subfigure}
\usepackage{amssymb}
\usepackage{array}
\title{Domain Independent SVM for Transfer Learning in Brain Decoding}

\author{
Shuo Zhou$^1$
\and
Wenwen Li$^1$\and
Christopher R. Cox$^2$\and
Haiping Lu$^{1,3}$
\affiliations
$^1$Department of Computer Science, University of Sheffield, United Kingdom\\
$^2$Department of Psychology, Louisiana State University, Baton Rouge, United States\\
$^3$Sheffield Institute for Translational Neuroscience, United Kingdom
\emails
\{szhou20, wenwen.li, h.lu\}@sheffield.ac.uk,
\{chriscox\}@lsu.edu
}

\begin{document}

\maketitle

\begin{abstract}
Brain imaging data are important in brain sciences yet expensive to obtain, with big volume (i.e., large $p$) but small sample size (i.e., small $n$). To tackle this problem, transfer learning is a promising direction that leverages source data to improve performance on related, target data. Most transfer learning methods focus on minimizing data distribution mismatch. However, a big challenge in brain imaging is  the large domain discrepancies in cognitive experiment designs and subject-specific structures and functions. A recent transfer learning approach minimizes domain dependence to learn common features across domains, via the \textit{Hilbert-Schmidt Independence Criterion} (HSIC). Inspired by this method, we propose a new \textit{Domain Independent Support Vector Machine} (DI-SVM) for transfer learning in brain condition decoding. Specifically, DI-SVM simultaneously minimizes the SVM empirical risk and the dependence on domain information via a simplified HSIC. We use public data to construct 13 transfer learning tasks in brain decoding, including three interesting multi-source transfer tasks. Experiments show that DI-SVM's superior performance over eight competing methods on these tasks, particularly an improvement of more than 24\% on multi-source transfer tasks.

\end{abstract}

\section{Introduction}
High resolution imaging is playing a major role in progressing science in many basic and applied areas.
In cognitive neuroscience, neuroimaging is used to relate different cognitive functions to patterns of neural activity supported by complex networks of interacting brain regions.
This often takes the form of a classification problem \citep{Singh2007DetectionOC}.
Machine learning techniques have been applied to sets of neuroimages to distinguish between \textit{brain conditions} associated with experimental manipulations.
However, while functional magnetic resonance imaging (fMRI; \cite{ogawa1990brain}) produces volumes with hundreds of thousands of potentially relevant voxels, a typical experiment will have on the order of 100 discrete trials.
This severely constrains the number of training examples available for the classifier.

Transfer learning is a machine learning scheme that can improve the classification performance on a learning task by leveraging the knowledge from related tasks \citep{pan2010survey}.
The task of interest is called the \textit{target domain}, while the task(s) to be leveraged is called the \textit{source domain} \citep{pan2010survey}.
A transfer learning problem is \textit{homogeneous} when the feature and label space of the source and target domains are the same, and \textit{heterogeneous} if they are different \citep{weiss2016survey}.

For the brain decoding problem mentioned above, there are public brain imaging data from multiple sites available, e.g., the OpenNeuro \citep{gorgolewski2017openneuro} and human connectome project (HCP) \citep{van2012human}.
For these data, there are many overlapped/similar brain conditions across different cognitive experiments, making homogeneous transfer a promising solution. 
However, existing transfer learning studies on brain decoding \citep{mensch2017learning,zhang2018transfer} did not consider aggregating data from multiple experiments with the same/similar labels when training a decoding model.
Here, we make the first attempt, to the best of our knowledge, to investigate \textit{homogeneous transfer learning for brain condition decoding}.
\par
Homogeneous transfer learning methods are mainly studied in computer vision (CV) and natural language processing (NLP). They focus on minimising data distribution mismatch, i.e., making features from the source and target to have as similar distributions as possible. Two approaches are popular: 1) learning a feature mapping to minimize the distribution mismatch \citep{pan2011domain,long2013jda}; and 2) jointly optimizing the mapping and classifier parameters \citep{xiao2015feature,chu2017selective,cao2018unsupervised,wang2018visual}.
\par
However, decoding brain conditions presents challenges different from those in CV/NLP. The same experiment paradigm can have different designs across sites.
Differences in individual brain structures and functions make informative voxels to vary across subjects \citep{NIPS2015_5855}.
Therefore, some methods in neuroscience consider each subject as a learning task to extract subject-specific features \citep{rao2013sparse}.
Transfer learning for brain condition decoding needs to take differences in domain information, such as experiment designs and subjects, into account. 

\textit{Maximum independence domain adaptation} (MIDA) \citep{yan2018learning} introduces a new domain dependence minimization approach to transfer learning. It constructs auxiliary features to encode domain information and learns common cross-domain features with minimum dependence on the domain information, as measured by the Hilbert-Schmidt Independence Criterion (HSIC) \citep{gretton2005measuring}.
This inspired us to follow a similar approach for brain condition decoding problems, where we can encode different experiment designs and subjects as auxiliary domain information.

In this paper, we propose \textit{Domain Independent Support Vector Machine} (DI-SVM), a homogeneous transfer classifier for brain condition decoding. It is build on MIDA, which shows that a classifier depending less on domain information can better leverage source data to improve target data prediction. Therefore, we aim to learn a generalized decoding model with minimum dependency on domain information. Specifically, DI-SVM simultaneously minimizes the Support Vector Machine (SVM) empirical risk and the dependence on experiment designs and subjects via a simplified HSIC. This is our first contribution.

Our second contribution is to construct a set of transfer learning problems for brain condition decoding studies. We identify subsets with the same or similar brain conditions from public brain imaging repositories. With the help of a neuroscientist, we carefully design six transfer learning cases (with 13 transfer learning tasks) of increasing difficulties from a psychological perspective. Experimental results show that DI-SVM  outperforms six state-of-the-art transfer learning methods on these tasks, in particular more than \textbf{24\%} improvement on multi-source transfer tasks. This confirms that homogeneous transfer learning is a promising direction for brain condition decoding, and brain sciences more broadly.
\label{sec:intro}

\section{Preliminaries}
\label{sec:preliminary}
Before introducing DI-SVM, we review the HSIC for measuring statistical dependence and three approaches in state-of-the-art transfer learning methods.

\textbf{Hilbert-Schmidt Independence Criterion (HSIC)} is a non-parametric criterion for measuring the statistical dependence between two sets $ \mathbf{X}=\{\mathbf{x}_i\}  $ and $ \mathbf{Y}=\{\mathbf{y}_i\} $, both with $n$ samples.
HSIC is zero if and only if $ \mathbf{X}$ and $ \mathbf{Y} $ are independent. A larger HSIC value suggests stronger dependence.
The empirical HSIC between $\mathbf{X}$ and $\mathbf{Y}$, $\rho_h(\mathbf{X, Y})$, is given by \citep{gretton2005measuring}
\begin{equation}
\rho_h(\mathbf{X, Y}) = \frac{1}{(n-1)^{2}} \text{tr}(\mathbf{KHLH}),
\label{eq:hsic}
\end{equation}
where $ \mathbf{K, H, L} \in \mathbb{R}^{n\times n} $, $ \mathbf{K}_{i,j} := k_x(\mathbf{x}_i, \mathbf{x}_j) $, $ \mathbf{L}_{i,j} := k_y(\mathbf{y}_i, \mathbf{y}_j) $, $k_x(\cdot)$, $k_y(\cdot)$ are two kernel functions, e.g., linear, polynomial, or radial basis function (RBF), $ \mathbf{H} = \mathbf{I} -\frac{1}{n}\mathbf{11^\top} $ is the centering matrix, and $\text{tr}\left(\cdot\right)$ is the trace function.
 
\textbf{Distribution Mismatch Minimization Mapping.}
The most popular approach to homogeneous transfer learning is to minimize the distribution mismatch between a source and a target, typically measured by the \textit{maximum mean discrepancy} (MMD) criterion \citep{borgwardt2006integrating}, via learning a mapping of the input data to a subspace. \cite{pan2011domain} proposed \textit{Transfer Component Analysis} (TCA) to learn such a mapping by minimizing the MMD of marginal distribution mismatch and maximizing the captured variances. \cite{pan2011domain} also proposed \textit{Semi-Supervised TCA} (SSTCA) for extracting more discriminate features by maximizing HSIC on the data labels.
\citet{long2013jda} extended TCA to \textit{Joint Distribution Adaptation} (JDA) to minimize both marginal and conditional distribution mismatch. We can summarize these methods in a general formula
\begin{equation}
\min_{\phi} \mathcal{D}(\phi(\mathbf{X}_s), \phi(\mathbf{X}_t)) - \mathrm{Var}(\phi(\mathbf{X})),
\label{eq:feat_dist_ali}
\end{equation}
where $ \phi $ denotes a feature mapping, $ \mathbf{X}_s $ and $ \mathbf{X}_t $ are source and target domain data, respectively, $ \mathcal{D}(\phi(\mathbf{X}_s), \phi(\mathbf{X}_t)) $ is the distribution mismatch (MMD) between source and target domains, $ \mathbf{X} $ is the combined source and target data, $ \mathrm{Var}(\cdot) $ denotes the total captured variance. A regularization term on the feature mapping matrix is typically incorporated.
SSTCA has a label dependence objective $ -\mu \rho_h(\phi(\mathbf{X}), \mathbf{Y}) $ added to Eq. (\ref{eq:feat_dist_ali}), where $ \mu$ is a hyper-parameter, and $ \mathbf{Y} $ is a label matrix (e.g. $y_{i,j} = 1$ if $\mathbf{x}_i$ belongs to the $j$th class; $y_{i,j} = 0$ otherwise).

\begin{table}[t]
	\setlength{\tabcolsep}{1pt}
	\centering
	\begin{tabular}{ l  l  m{4.56cm} }
		\toprule
		\textbf{Method$\:\:$} & $ \mathcal{L}(f(\phi(\mathbf{X}^l), \mathbf{y}))\:\: $ & $ \mathcal{D}(\phi(\mathbf{X}_s), \phi(\mathbf{X}_t)) $   \\
		\midrule
		SSMDA  & Square loss & $ -\rho_h(\phi(\mathbf{X}_s), \phi(\mathbf{X}_t)) $ \\ 
		STM  & Hinge loss & MMD of marginal dist mismatch\\ 
		DMM & Hinge loss & MMD of marginal dist mismatch \\ 
		MEDA & Square loss & MMD of marginal and conditional dist mismatch\\
		\bottomrule	
	\end{tabular}
	\caption{The four domain-invariant classifier learning methods as in Eq. (\ref{eq:diclf}): \textit{Semi-supervised kernel matching domain adaptation} (SSMDA) \citep{xiao2015feature}, \textit{Selective Transfer Machine} (STM) \citep{chu2017selective}, \textit{Distribution Matching Machine (DMM)} \citep{cao2018unsupervised}, and \textit{Manifold Embedded Distribution Alignment} (MEDA) \citep{wang2018visual}. ``dist'' in the table denotes distribution. }
	\label{tab:tl_clf}
\end{table}

\textbf{Domain-Invariant Classifier Learning.} Another approach learns a classifier by minimizing the prediction error and distribution mismatch jointly. We summarize this approach using the following formula
\begin{equation}
\min_{f, \phi} \mathcal{L}(f(\phi(\mathbf{X}^l), \mathbf{y})) + \lambda \mathcal{D}(\phi(\mathbf{X}_s), \phi(\mathbf{X}_t)),
\label{eq:diclf}
\end{equation}
where $ f(\cdot) $ is a decision function of a classifier, $\mathcal{L}(f(\phi(\mathbf{X}^l), \mathbf{y}))$ is the prediction error, $ \mathbf{X}^l $ denotes labeled data, which can be source data only or source data plus target data according to different settings, and $ \mathbf{y} $ is a data label vector. Table \ref{tab:tl_clf} summarizes four such recent methods. 

\textbf{Domain Dependence Minimization Mapping.}
\citet{yan2018learning} proposed the MIDA method using a new approach. It obtains cross-domain features by learning a mapping to minimize the dependence on auxiliary domain information. In contrast, such domain information is not directly modeled in the distribution mismatch minimization mapping or domain-invariant classifier learning approaches.
We summarize the objective of MIDA as
\begin{equation}
\min_{\phi}  \rho_h(\phi(\mathbf{X}), \mathbf{A}) - \textrm{Var}(\phi(\mathbf{X})),
\label{eq:mida}
\end{equation}
where $ \mathbf{A} $ encodes the auxiliary domain information. Similar to TCA, MIDA also has a semi-supervised version (SMIDA) that maximizes label dependence by adding a term $ -\mu \rho_h(\phi(\mathbf{X}), \mathbf{Y}) $ to Eq. (\ref{eq:mida}).

\section{Domain Independent SVM for Brain Decoding}
\label{sec:DISVM}
In this section, we develop DI-SVM, a new transfer learning method for the brain condition decoding problem. We will first describe the transfer learning setting for this problem and then propose the DI-SVM model and algorithm, followed by some discussions.

\subsection{Transfer Learning Setting for Brain Decoding}
\label{subsec:prob_form}
We aim to learn a generalized brain condition decoding model from source and target data samples while utilizing the domain information. We consider a \textbf{semi-supervised} setting where all source data samples are labeled while the target data have both labeled samples and unlabeled samples. The objective is to predict the labels of \textit{unlabeled target samples}.

\textbf{The target.} The target cognitive experiment has $ n_t $ fMRI data samples $\mathbf{X}_t \in \mathbb{R}^{d \times n_t}$ of $ m $ brain conditions, where $ d $ is the dimension of an fMRI data sample, i.e., the number of raw fMRI features. There are $\tilde{n}_t$ labeled samples and (${n}_t - \tilde{n}_t $) unlabeled samples in the target domain. 

\textbf{The source.} The source consists of data from one or more cognitive experiments with $n_s$ labeled fMRI data samples $\mathbf{X}_s \in \mathbb{R}^{d \times n_s}$ in total, with the same $m$ brain conditions as the target data. 

\textbf{Domain information encoding.} Denote the target and source data jointly as $ \mathbf{X} = [\mathbf{X}_s,\mathbf{X}_t] \in \mathbb{R}^{d \times n} $, $ n = n_s + n_t $. Each fMRI sample $ \mathbf{x}_i $ ($i=1,\cdots,n$) is collected with a particular experiment design $j$ from a particular subject $k$, where $j=1,\cdots,p$ and $k=1,\cdots,q$, i.e., there are $p$ unique experiment designs and $q$ unique subjects. This is the \textit{domain information} to be utilized in our transfer learning method, following MIDA \citep{yan2018learning} as reviewed above. We use a simple \textit{one-hot-encoding} strategy to encode such domain information. Specifically, we construct an experiment design one-hot matrix $ \mathbf{E} \in \mathbb{R}^{n\times p} $, where its ($i,j$)th element $ e_{i,j} = 1 \text{ if } \mathbf{x}_i $ is collected from experiment $ j $ and $ e_{i,j} = 0 $ otherwise.
Similarly, we construct a subject one-hot matrix $ \mathbf{S}  \in \mathbb{R}^{n\times q} $, where $ s_{i,k} = 1 \text{ if } \mathbf{x}_i $ is from subject $ k $ and $ s_{i,k} = 0 $  otherwise. We then obtain the auxiliary domain information matrix $ \mathbf{A} \in \mathbb{R}^{\hat{d}\times n}$ by concatenating $ \mathbf{E}^\top $ and $ \mathbf{S}^\top $, where $ \hat{d} = p+q $. 

\subsection{Domain Independent SVM (DI-SVM)}
\label{subsec:model}
\textbf{Domain-independent classifier learning}. Domain-invariant classifier learning methods (Eq. (\ref{eq:diclf})) minimize prediction errors directly to optimize prediction performance, while domain dependence minimization mapping allows us to leverage important domain information such as the experiment designs and subjects for transfer learning. Here we propose the DI-SVM model by combining the virtues from both approaches, i.e., minimizing prediction error and domain dependence simultaneously. Specifically, we can replace the distribution mismatch term in Eq. (\ref{eq:diclf}) with a domain dependence term using HSIC in Eq. (\ref{eq:mida}). This gives the following domain-independent classifier learning model as 
\begin{equation}
    \min_{f} \mathcal{L}(f(\phi(\mathbf{X}^l), \mathbf{y})) + \lambda \rho_h(\phi(\mathbf{X}), \mathbf{A}),
    \label{eq:model}
\end{equation}
where $ \lambda>0 $, $ \mathbf{X}^l \in \mathbb{R}^{d \times\tilde{n}} $ denotes all labeled samples from source and target domains, $ \tilde{n} = \tilde{n}_t + n_s$, and $ \mathbf{y} $ is the label vector. In this paper, we consider only \textit{binary} labels, i.e., $y_i \in \{-1, 1\} $, $i=1,\cdots \tilde{n}$. Accordingly, we consider only $m=2$, i.e., the classification of two brain conditions.

\textbf{SVM classifier.}
We can plug in any loss function for the first term in our model Eq. (\ref{eq:model}), such as the square loss, logistic loss, or hinge loss. 
Here we choose the hinge loss for empirical risk minimization in SVM. The linear SVM has a target decision function in a simple form $ f(\mathbf{x}) = \boldsymbol{\beta}^\top \mathbf{x} $, where $ \boldsymbol{\beta}$ is a coefficient vector. The standard linear SVM learns the decision function by solving the following minimization problem
\begin{equation}
\begin{gathered}
    \min_{\boldsymbol{\beta, \xi}} \frac{1}{2}\|\boldsymbol{\beta}\|^2+C\sum_i^{\tilde{n}} \xi_i,\\
    \mathrm{s.t.}\:\:  y_i \boldsymbol{\beta}^\top\mathbf{x}_i\leq1-\xi_i, y_i \in \{-1, 1\},
    \label{eq:linear_SVM}
\end{gathered}
\end{equation}
where $\xi_i$ is the ``slack variable'' for the $i$th sample, $ C $ is a hyper-parameter controlling the trade-off between more accurate model on training data and larger margin of decision hyperplane, and $\tilde{n}$ is the number of labeled sample. The linear SVM can be extended to a nonlinear version by introducing a kernel function $ k(\cdot) $. Using the Representer Theorem \citep{scholkopf2001generalized}, the $ i $th sample can be represented as $\mathbf{k}_i = k(\mathbf{x}_i,\mathbf{X})$, yielding the Kernel SVM
\begin{equation}
\begin{gathered}
    \min_{\boldsymbol{\beta, \xi}} \frac{1}{2}\boldsymbol{\beta}^\top\mathbf{K}\boldsymbol{\beta}+C\sum_i^{\tilde{n}}\xi_i,\\
    \mathrm{s.t.}\:\:  y_i \boldsymbol{\beta}^\top\mathbf{k}_i\leq1-\xi_i, y_i \in \{-1, 1\},
    \label{eq:kSVM}
\end{gathered}
\end{equation}
where $\mathbf{K} = k(\mathbf{X,X})$.

\textbf{Simplified HSIC.} 
With the above SVM-based formulation, we can construct a simplified HSIC by viewing the coefficient vector $\boldsymbol{\beta}$ as a \textit{classifier-based feature mapping}. This mapping projects input features to a one-dimensional space (i.e., a line), where the projected values represent the label likelihood. Following the principle of dependence minimization, we aim to learn a domain-independent classifier by minimizing the dependence of the label likelihood (projected values) on domain information, i.e., experiment designs and subjects. Therefore, we simplify the HSIC $\rho_h(\phi(\mathbf{X}), \mathbf{A})$ to the following version
\begin{align}
\begin{aligned}
\rho_{sh}(\phi(\mathbf{X}), \mathbf{A})& = \text{tr}((\boldsymbol{\beta}^\top \mathbf{K})^\top(\boldsymbol{\beta}^\top \mathbf{K})\mathbf{HK}_a\mathbf{H})\\ &= \boldsymbol{\beta}^\top\mathbf{K HK}_{a}\mathbf{HK}\boldsymbol{\beta},
\end{aligned}
\label{eq:HSIC_DISVM}
\end{align}
where $ \mathbf{K}_{a} = k_{a}(\mathbf{A, A})\in \mathbb{R}^{n \times n}$, and $ k_{a}(\cdot) $ is a kernel function. Due to the simplicity of $\mathbf{A}$ (with one-hot encoding), we always use a simple linear kernel here. Our simplified HSIC is constructed directly from classifier coefficient so there is no separate feature mapping step as in STM \citep{chu2017selective} and DMM \citep{cao2018unsupervised}. 

\textbf{Label recoding.}
In a semi-supervised setting of a binary classification problem, a sample $ \mathbf{x}_i $ can have a positive label `1', a negative label `-1', or no label, which is denoted as label `0'. Therefore, we recode the label vector $ \mathbf{y}$ into another label vector $ \tilde{\mathbf{y}}$, where $ \tilde{y}_i \in \{-1, 1\} $ if the $ i $th sample is labeled, $ \tilde{y}_i =0  $ otherwise, $i=1,\cdots,n$.

\textbf{DI-SVM model and algorithm.}
Using Eqs. (\ref{eq:kSVM}) and (\ref{eq:HSIC_DISVM}), we can formulate the objective function of DI-SVM  as
\begin{equation}
\begin{gathered}
    \min_{\boldsymbol{\beta, \xi}} \frac{1}{2}\boldsymbol{\beta}^\top\mathbf{K}\boldsymbol{\beta} + C\sum_i^{n}\xi_i +  \frac{\lambda}{2}\boldsymbol{\beta}^\top\mathbf{K HK}_{a}\mathbf{HK}\boldsymbol{\beta},\\
    \mathrm{s.t.}\:\:  \tilde{y}_i \boldsymbol{\beta}^\top\mathbf{k}_i\leq1-\xi_i, \tilde{y}_i \in \{-1, 1, 0\}.
\end{gathered}
\label{eq:ovr_obj}
\end{equation}
This is a quadratic programming (QP) problem that can be solved by standard QP tools.  Algorithm \ref{alg:DISVM} is the pseudocode of DI-SVM.

\textbf{The role of unlabeled samples.} DI-SVM requires all samples, both labeled and unlabeled, in training. Unlabeled samples can only influence $\boldsymbol{\beta} $ in the simplified HSIC term, i.e. dependence on domain information. This is because when the $ i $th sample is unlabeled, i.e., $ \tilde{y}_i=0 $, we have $ \tilde{y}_i \boldsymbol{\beta}^\top\mathbf{k}_i = 0 $, which means that this sample has no influence on $ \boldsymbol{\beta} $ in the first term of SVM empirical risk.

\textbf{Relationship with SMIDA.} SMIDA is the closest existing model to the proposed DI-SVM. We can view DI-SVM as 1) replacing the label dependence term $  \rho_h(\phi(\mathbf{X}), \mathbf{Y}) $ in SMIDA with an SVM empirical risk term (hinge loss function), and 2) learning a mapping to a \textit{one-dimensional} classification space (i.e., a line) rather than a low-dimensional subspace.

\begin{algorithm}[tb]
	\caption{Domain Independent SVM (DI-SVM)}
	\label{alg:DISVM}
	\textbf{Input}: Input feature matrix $ \mathbf{X}= [\mathbf{X}_s,\mathbf{X}_t] \in \mathbb{R}^{d \times n}  $, auxiliary domain information, label vector $ \mathbf{y}$.\\
	\textbf{Hyper-parameter}: Penalty $C$ for ``slack variable'', regularization parameter $ \lambda $ for the domain independence term,  and kernel hyper-parameter(s) when using non-linear kernels.\\
	\textbf{Output}: Coefficient vector $ \boldsymbol{\beta} $. 
	
	\begin{algorithmic}[1] 
		\STATE Encode auxiliary domain information into a matrix $ \mathbf{A} \in \mathbb{R}^{\hat{d}\times n}$ with one-hot-encoding, see Section \ref{subsec:prob_form};
		
		\STATE Recode label vector $ \mathbf{y}$ into $ \tilde{\mathbf{y}} $, see Section \ref{subsec:model};
		
		\STATE Construct kernel matrices $ \mathbf{K} = \phi(\mathbf{X})^\top\phi(\mathbf{X})$, $ \mathbf{K}_{a} =  \mathbf{A}^\top\mathbf{A}$,  and centering matrix $ \mathbf{H} $;
		
		\STATE Solve QP problem  for Eq. (\ref{eq:ovr_obj});
		\STATE \textbf{return} Coefficient vector $ \boldsymbol{\beta} $.
	\end{algorithmic}
\end{algorithm}

\section{Experiments}
To evaluate DI-SVM, we design six transfer learning cases with 13 transfer learning tasks for brain decoding, built from six datasets, and compare DI-SVM against eight competing methods. 


\begin{table}[t]
	\setlength{\tabcolsep}{2pt}
	\centering
	\begin{tabular}{ c  c  m{5cm}  c }
		\toprule
		Exp &\#AC & Exp Description & \#Samples  \\
		\midrule
		A & ds007 & Stop signal with spoken pseudo word naming \citep{xue2008common} & 39 \\ 
		B & ds007 &  Stop signal with spoken letter naming \citep{xue2008common} & 38 \\
		C & ds007 & Stop signal with manual response \citep{xue2008common} & 40 \\
		D & ds008 & Conditional stop signal \citep{aron2007triangulating} & 26 \\
		E & ds101 & Simon task [Unpublished] & 42 \\
		F & ds102 & Flanker task \citep{kelly2008competition} & 52 \\
		\bottomrule	
	\end{tabular}
	\caption{Information on the OpenfMRI data used. `Exp' indexes the six cognitive experiments A--F. \#AC is the accession number of an OpenfMRI project, where the same group of subjects are used in each project and there is no overlapping subject between projects. Each of the six experiments has two brain conditions to classify and \#Sample indicates the number of samples for each brain condition.}
	\label{tab:data}
\end{table}


\textbf{Dataset selection.} With the help of a neuroscientist, we carefully selected six datasets (A to F) that are most meaningful from psychological perspective from the public OpenfMRI repository.\footnote{\url{https://legacy.openfmri.org} or \url{https://openneuro.org}.} Each dataset is from an experiment, which is treated as a \textbf{domain}. Table \ref{tab:data} summarizes basic information on these datasets. Each dataset contains fMRI time-series data for multiple subjects. Subjects from the same accession number (ds$\times\!\!\times\!\!\times$) are the same and there is no overlapped subject between accession numbers. There are two brain conditions selected from each dataset. Each is considered as a \textit{class} and has the same number of samples. Thus, we have binary classification problems that discriminate between brain conditions in an experiment.

\textbf{Preprocessing.} Each sample was preprocessed with the protocol published by \citet{poldrack2013toward} to obtain the Z-score statistical parametric map (SPM) of size $ 91 \times 109 \times 91 $, which is then reduced to a vector of size $ 228,546 $ containing only the valid voxels (some are outside of the brain).

\begin{table}[t]
	\setlength{\tabcolsep}{3pt}
	\centering
	\begin{tabular}{ c  c  m{6cm} }
		\toprule
		Case & Exp & Pos vs Neg Conditions for Classification \\
		\midrule
		\multirow{2}{*}{1} & A & Successful stop vs Unsuccessful stop \\\cline{2-3}
		& B & Successful stop vs Unsuccessful stop \\
		\midrule
		\multirow{2}{*}{2} & A & Successful stop vs Unsuccessful stop\\\cline{2-3}
		& C & Successful stop vs Unsuccessful stop \\\midrule
		\multirow{2}{*}{3} & B & Successful stop vs Unsuccessful stop	\\\cline{2-3}
		& C & Successful stop vs Unsuccessful stop \\
		\midrule
		\multirow{2}{*}{4} & A & Successful stop vs Unsuccessful stop \\\cline{2-3}
		& D & Successful stop - critical vs Unsuccessful stop - critical \\\midrule
		\multirow{2}{*}{5} & E & Congruent correct vs Incongruent correct  \\\cline{2-3}
		& F & Congruent correct vs Incongruent correct\\
		\midrule
		\multirow{3}{*}{6} & A & Successful stop vs Unsuccessful stop \\\cline{2-3}
		& B & Successful stop vs Unsuccessful stop \\\cline{2-3}
		& C & Successful stop vs Unsuccessful stop \\
		\bottomrule	
	\end{tabular}
	\caption{Transfer learning cases formulated for experimental evaluation. `Case' denotes the transfer cases as described in text. `Exp' denotes the cognitive experiments. Cases 1 to 5 are for single-source transfer and Case 6 is for multi-source transfer.}
	\label{tab:transfer_task}
\end{table}

\textbf{Six transfer learning cases.} With the help of a neuroscientist, we constructed six distinct cases of transfer learning problems with increasing difficulties, according to the paradigms used (whole experimental processes), subjects involved, cognitive \textit{control} demands, and response types. Table \ref{tab:transfer_task} summarizes these six cases. 

\textbf{Case 1: Same paradigm, subject, and control, different responses.}
The first case aims to transfer between two \textit{stop signal tasks} completed by the same subjects.
These tasks involve producing conditioned responses to simple stimuli, but inhibiting the response on 25\% of trials preceded by a stop signal.
The latency between the stop signal and stimulus onset determines the difficulty in inhibiting the response, and latency was adjusted during the experiment so that subjects failed to inhibit about 50\% of the time.
Both tasks required \textit{verbal responses} and involved the same cognitive control demands, but differed in the \textit{linguistic complexity of the response}: 1) the stimulus were strings of letters that were not English words but could be ``sounded out'' and the response was to pronounce each string as an English pseudo-word; 2) the stimulus were single letters and the response was to name them aloud. Successful and unsuccessful stops are the two brain conditions to classify.

\textbf{Cases 2 \& 3: Same paradigm and subjects, different controls.}
Both cases aim to transfer between each of the stop signal tasks from Case 1 and a nearly identical task with the same subjects, but requiring \textit{a manual response} (button press with index or middle finger) rather than a \textit{verbal response}. Again, the two brain conditions to classify are successful and unsuccessful stops.

\textbf{Case 4: Same paradigm, different subjects and controls.}
This case studies the transfer between experiments A and D, which were conducted on different subjects. The primary cognitive difference is that the conditional stop signal in experiment D should only be heeded for one of the two stimuli.
Thus, these tasks differed in their cognitive control demands and response complexity, yet share the same general ``stop signal'' paradigm.
This case classified successful against unsuccessful stops as well.

\textbf{Case 5: Different paradigms and subjects, similar controls.} 
This case studies the transfer between the \textit{flanker task} (E) and the \textit{Simon task} (F).
They are distinct paradigms that differ in how stimulus are presented, the kinds of stimuli involved, and the source of task interference to be inhibited.
Yet, they share abstract similarity: they both involve overcoming a prevalent response bias and are understood to involve similar executive cognitive functions. The two brain conditions to classify correspond to congruent and incongruent trials.

\textbf{Case 6: Multi-source transfer with the same subjects.}
Cases 1--5 study transfer learning on increasingly difficult problems, where one source domain is leveraged to improve classification in a target domain.
This last case studies whether leveraging multiple source domains can obtain further improvement. From the three stop signal tasks introduced in Cases 1--3, we treat each possible pair as the source domain for the remaining task (as the target domain).

In total, there are 13 transfer learning tasks constructed.

\textbf{Eight methods compared.} We evaluate DI-SVM against eight methods: two simple baselines 1) \textbf{PCA}$_t$, where PCA was only performed on the target data, 2) \textbf{PCA}$_{s+t}$, where PCA was performed on the combined source and target data; and six state-of-the-art transfer learning methods discussed in Section \ref{sec:preliminary}: 3) \textbf{TCA}, 4) \textbf{SSTCA}, 5) \textbf{JDA}, 6) \textbf{MEDA}, 7) \textbf{MIDA}, and 8) \textbf{SMIDA}.
PCA, TCA, SSTCA, JDA, MIDA, and SMIDA only learn a feature mapping and they use SVM as the classifier. Both linear and RBF kernels were studied for such SVM classifiers, and also MEDA and DI-SVM. 
We will report the accuracy obtained by DI-SVM with both linear and RBF kernels. 
For other methods, we report the accuracy from the best performing kernel. 

\textbf{Experimental Settings.}
We performed 10 $ \times $ 5-fold cross-validation on the target domain. For each split, the target training samples and all source samples (except for PCA$_t$) were used for training, and the remaining target test samples are for testing. On each training set, we determined the best hyper-parameters via grid search with 20 further random splits (20\% for validation and 80\% for training).
For DI-SVM, we first fixed $ \lambda = 1 $ and searched for the best $C$ in $[10^{-3}, 10^4] $ on regular grids of log scale with a step size of one (i.e., eight values). Then we fixed the best $C$ and searched for the best $\lambda$ among five values only: $\{0.01, 0.1, 1, 10, 100\}$. Sensitivity studies will be provided in the end of this section.

We also tuned hyper-parameters for all the other methods.
For PCA, TCA, SSTCA, JDA, MIDA and SMIDA, we searched for the best subspace dimension from $ \{20, 40, 50, 60 ,80, 100\} $.  
For TCA, SSTCA, JDA, MIDA and SMIDA, linear kernel was used. 
For SVM, we searched for the best $ C \in [10^{-3}, 10^4] $, and the best Gamma value in $ [10^{-6}, 10^2] $ when using the RBF kernel.
For other method-specific hyper-parameters, we followed the strategies stated in their original papers.

We used one-hot encoding as presented in Section \ref{subsec:prob_form} to obtain the domain information matrix $ \mathbf{A} $ for DI-SVM, MIDA, and SMIDA. DI-SVM, SSTCA, and SMIDA also intake the label vector $ \tilde{\mathbf{y}} $ (see Section \ref{subsec:model}). Linear kernel was used to construct the label kernel matrix for SSTCA and SMIDA, i.e. $ \mathbf{K}_y = \mathbf{\tilde{y}^\top \tilde{y}}$.

\textbf{Experimental Results and Discussions.}
\label{subsec:results}
\begin{table*}[t]
	\small
	\setlength{\tabcolsep}{4pt}
	\centering
	\begin{tabular}{ m{1.3cm}  c  c  c  c  c  c  c  c  c  c  c }
		\toprule
		& \multicolumn{2}{c}{Case 1}  &  \multicolumn{2}{c}{Case 2} & \multicolumn{2}{c}{Case 3} & \multicolumn{2}{c}{Case 4} & \multicolumn{2}{c}{Case 5} & \multirow{2}{0.8cm}{Ave}\\
		& A$ \to $B & B$ \to $A & A$ \to $C & C$ \to $A & B$ \to $C & C$ \to $B & A$ \to $D & D$ \to $A & E$ \to $F & F$ \to $E& \\
		\midrule
		PCA$_t $ & 58.9$ \pm $4.6 & 57.8$ \pm $4.0 & 65.5$ \pm $4.6 & 57.8$ \pm $4.0 & 65.5$ \pm $4.6 & 58.9$ \pm $4.6 & 79.6$ \pm $8.2 & 57.8$ \pm $4.0 & \underline{67.5$ \pm $6.2} & 51.8$ \pm $3.9 & 62.1\\
		PCA$_{s+t} $ & 59.7$ \pm $4.5 & 60.8$ \pm $3.1 & 74.0$ \pm $5.9 & 67.6$ \pm $5.2 & 78.9$ \pm $3.4 & 66.6$ \pm $4.0 & 51.9$ \pm $4.3 & \underline{63.7$ \pm $5.3} & 51.4$ \pm $1.8 & 51.3$ \pm $2.4 & 62.6\\
		\midrule
		TCA & 55.0$ \pm $4.3 & 58.9$ \pm $3.2 & 70.6$ \pm $6.3 & 63.6$ \pm $4.0 & \underline{86.4$ \pm $5.3} & \underline{75.3$ \pm $4.0} & 54.0$ \pm $3.2 & 58.0$ \pm $4.8 & 52.3$ \pm $3.2 & 49.3$ \pm $1.9 & 62.3 \\
		SSTCA & 48.0$ \pm $3.9 & 50.4$ \pm $4.6 & 48.0$ \pm $4.6 & 51.8$ \pm $3.4 & 49.9$ \pm $3.6 & 52.6$ \pm $4.4 & 53.7$ \pm $3.2 & 50.0$ \pm $3.1 & 56.0$ \pm $2.3 & 52.1$ \pm $2.1 & 51.2 \\		
		JDA	& 63.7$ \pm $2.8 & 54.4$ \pm $1.8 & 58.4$ \pm $3.1 & 53.7$ \pm $3.6 & 63.1$ \pm $5.2 & 54.1$ \pm $3.5 & 64.2$ \pm $9.9 & 55.5$ \pm $1.9 & 50.5$ \pm $2.2 & 50.2$ \pm $1.7 & 56.8 \\
		MEDA & 63.2$ \pm $0.3 & 58.8$ \pm $1.0 & 56.6$ \pm $1.2 & 61.5$ \pm $0.5 & 64.3$ \pm $1.8 & 57.9$ \pm $0.3 & 53.4$ \pm $2.2 & 47.4$ \pm $0.3 & 52.4$ \pm $0.6 & \underline{54.3$ \pm $1.0} & 58.1 \\
		MIDA & \underline{64.5$ \pm $5.0}	& \underline{71.2$ \pm $3.5} & \underline{78.4$ \pm $2.9} & \underline{70.6$ \pm $4.0} & 80.1$ \pm $4.4 & 73.2$ \pm $3.3 & 63.3$ \pm $6.0 & 50.6$ \pm $2.4 & 66.4$ \pm $3.0 & 53.7$ \pm $4.5 & \underline{67.2} \\
		SMIDA & 54.9$ \pm $2.8 & 52.4$ \pm $1.5 & 57.5$ \pm $2.1 & 52.4$ \pm $1.3 & 58.5$ \pm $2.5 & 58.4$ \pm $2.4 & 72.7$ \pm $4.6 & 48.1$ \pm $1.7 & 61.5$ \pm $3.8 &	49.2$ \pm $1.7 & 56.6 \\

		\midrule
		$ \text{DI-SVM}_l $ & \textbf{79.0$ \pm $2.8} & \textbf{80.4$ \pm $3.5} & \textbf{86.8$ \pm $2.4} & \textbf{71.2$ \pm $2.6} & \textbf{92.1$ \pm $1.4} & \textbf{75.8$ \pm $1.8} & \textbf{87.1$ \pm $1.5} & \textbf{67.5$ \pm $4.1} & \textbf{68.9$ \pm $3.1} & 44.8$ \pm $4.9 & \textbf{75.3} \\
		$ \text{DI-SVM}_r $ & 61.5$ \pm $2.0 & 64.7$ \pm $2.8 & 73.8$ \pm $2.3 & 66.9$ \pm $2.4 & 69.4$ \pm $1.5 & 62.4$ \pm $2.8 & \underline{85.0$ \pm $2.2} & 60.6$ \pm $1.9 & 52.2$ \pm $0.8 & \textbf{55.7$ \pm $2.2} & 65.2 \\
		\bottomrule	
	\end{tabular}
	\caption{Classification accuracy in percentage for ten single-source transfer tasks (mean $ \pm $ standard deviation). `Ave' is the average over the ten tasks. Each transfer learning task is denoted as \textit{Source experiment}$\to$\textit{Target experiment}, e.g., A$ \to $B means A is the source and B is the target. Subscripts $ l $ and $ r $ denote linear and RBF kernels, respectively. The best results are in \textbf{bold} and the second best ones are \underline{underlined}.}
	\label{tab:results}
\end{table*}
\begin{table}[t]
	\centering
	\begin{tabular}{ m{1.4cm}  c  c  c  c }
		\toprule
		& B\&C$ \to $A & A\&C$ \to $B & A\&B$ \to $C & Ave\\
		\midrule
		PCA$_t $  & 57.8$ \pm $4.0 & 58.9$ \pm $4.6 & 65.5$ \pm $4.6 & 60.8 \\
		PCA$_{s+t} $ & 51.5$ \pm $1.5 & 53.7$ \pm $2.1 & 50.6$ \pm $2.5 & 52.0 \\\midrule
		TCA   & 52.1$ \pm $1.1 & 54.5$ \pm $3.1 & 52.1$ \pm $1.1 & 52.9 \\
		SSTCA & 52.1$ \pm $3.4 & 52.2$ \pm $3.2 & 56.5$ \pm $3.5 & 53.6 \\		
		JDA  & 51.7$ \pm $2.8 & 53.8$ \pm $3.3 & 52.0$ \pm $3.1 & 52.5 \\
		MEDA & 52.6$ \pm $1.4 & 52.6$ \pm $1.3 & 57.0$ \pm $0.8 & 54.1 \\
		MIDA & 51.4$ \pm $1.8 & 61.7$ \pm $2.2 & 65.4$ \pm $2.2 & 59.5 \\
		SMIDA & 55.4$ \pm $2.4 & 60.4$ \pm $2.2 & 60.5$ \pm $3.2 & 58.8 \\
		\midrule
		$ \text{DI-SVM}_l $ & \textbf{81.7$ \pm $1.8} & \textbf{81.3$ \pm $1.5} & \textbf{91.6$ \pm $2.2} & \textbf{84.9} \\
		$ \text{DI-SVM}_r $ & \underline{70.6$ \pm $1.5} & \underline{63.9$ \pm $1.9} & \underline{65.3$ \pm $1.7} & \underline{66.6} \\
		\bottomrule	
	\end{tabular}
	\caption{Classification accuracy in percentage for three multi-source transfer tasks. }
	\label{tab:multi_src_res}
\end{table}

Tables \ref{tab:results} and \ref{tab:multi_src_res} summarize the decoding accuracy of ten single-source transfer learning tasks and three multi-source transfer learning tasks respectively, with both the mean and standard deviation reported. The best results are highlighted in \textbf{bold} and the second best ones are \underline{underlined}. We have five observations:
\begin{itemize}
	\item DI-SVM outperformed all comparing methods, though the improvement decreased from easy to difficult cases. This indicates that more difficult cases from psychological perspective are also more difficult for transfer learning.
	\item For DI-SVM, multi-source transfer did not always outperform single-source transfer. The accuracy of transfer learning tasks, A\&C$\to$B and B\&C$\to$A are both higher than the corresponding single-source tasks A$\to$B and C$\to$B, and B$\to$A and C$\to$A, respectively. However, the accuracy of A\&B$\to$C is lower than that of  B$\to$C. This indicates that source selection can have large influence the performance. Nevertheless, DI-SVM outperformed all the other methods significantly (by more than \textbf{24\%} with the linear kernel) on the multi-source transfer tasks. Therefore, when there is no clear preference of a particular source, multi-source transfer is a preferred choice.
	\item HSIC-based methods MIDA and DI-SVM have outperformed other transfer learning methods in both single-source and multi-source transfer tasks. This shows the effectiveness of using domain information (via one-hot encoding) and performing dependence minimization on it (via HSIC). 
	\item PCA$_{s+t}$ outperformed PCA$_t $ in Cases 1 to 3, with all subjects common, but PCA$_t $ outperformed PCA$_{s+t}$ in Cases 4 and 5, with no subject in common.
	This demonstrates the large influence of domain information (subjects) on the classification accuracy. 
	\item In single-source transfer tasks, TCA and MIDA outperformed SSTCA and SMIDA, respectively, which may seem counter-intuitive because semi-supervised learning has more information (the labels) utilized. In contrast, in multi-source transfer tasks where there were more training samples, SSTCA slightly outperformed TCA on the whole, and the performance difference between SMIDA and MIDA was much smaller than in single-source transfer tasks. A possible explanation is that in small sample setting, maximizing dependence on training data labels is more susceptible to overfitting resulting poorer generalization performance. 
\end{itemize}

\textbf{Hyper-parameter Sensitivity.}
\begin{figure}[t]
	\centering
	\subfigure[$ C $\label{subfig:C}]{\label{fig:a}\includegraphics[width=41mm]{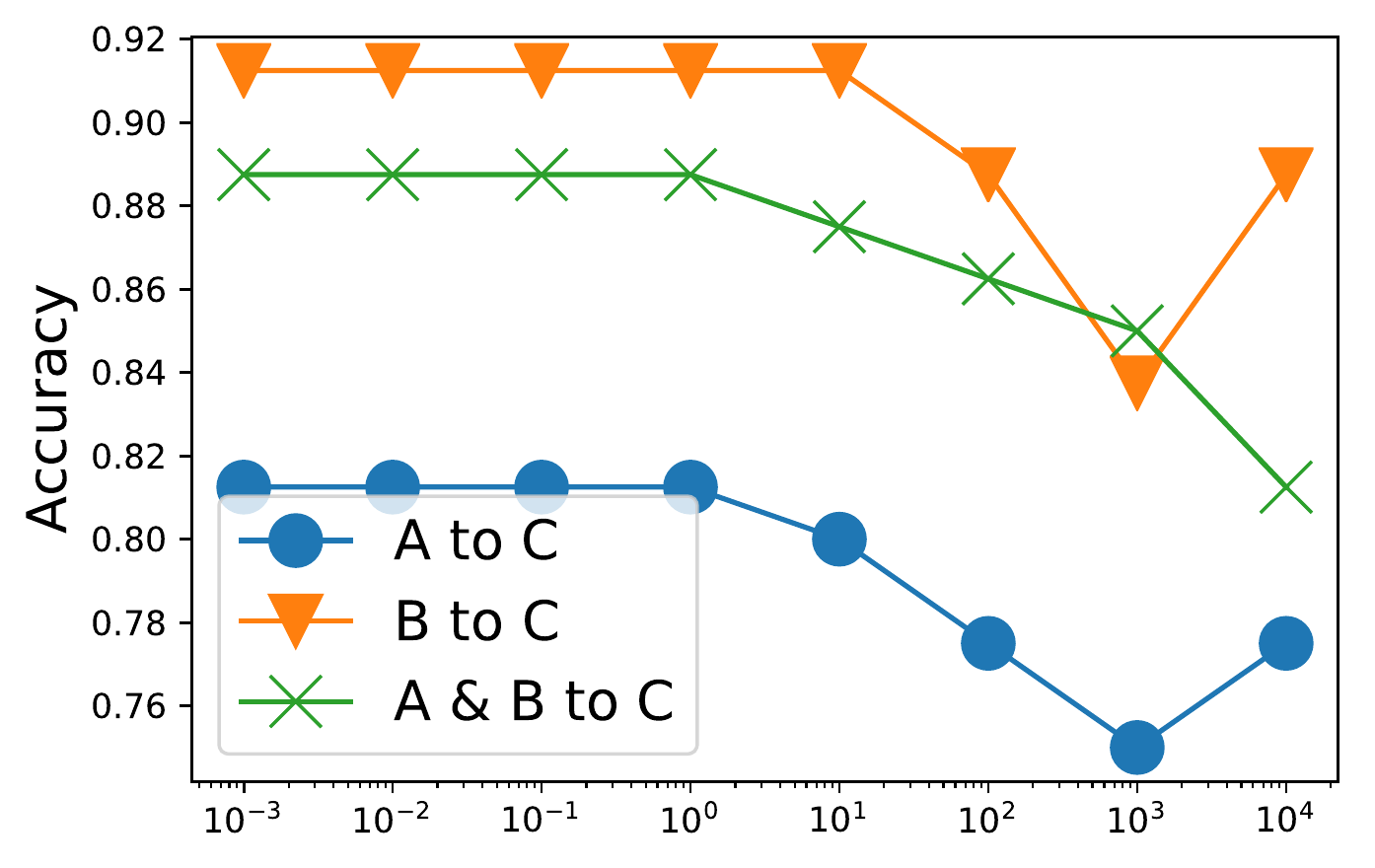}}
	\subfigure[$ \lambda $\label{subfig:lambda}]{\label{fig:b}\includegraphics[width=41mm]{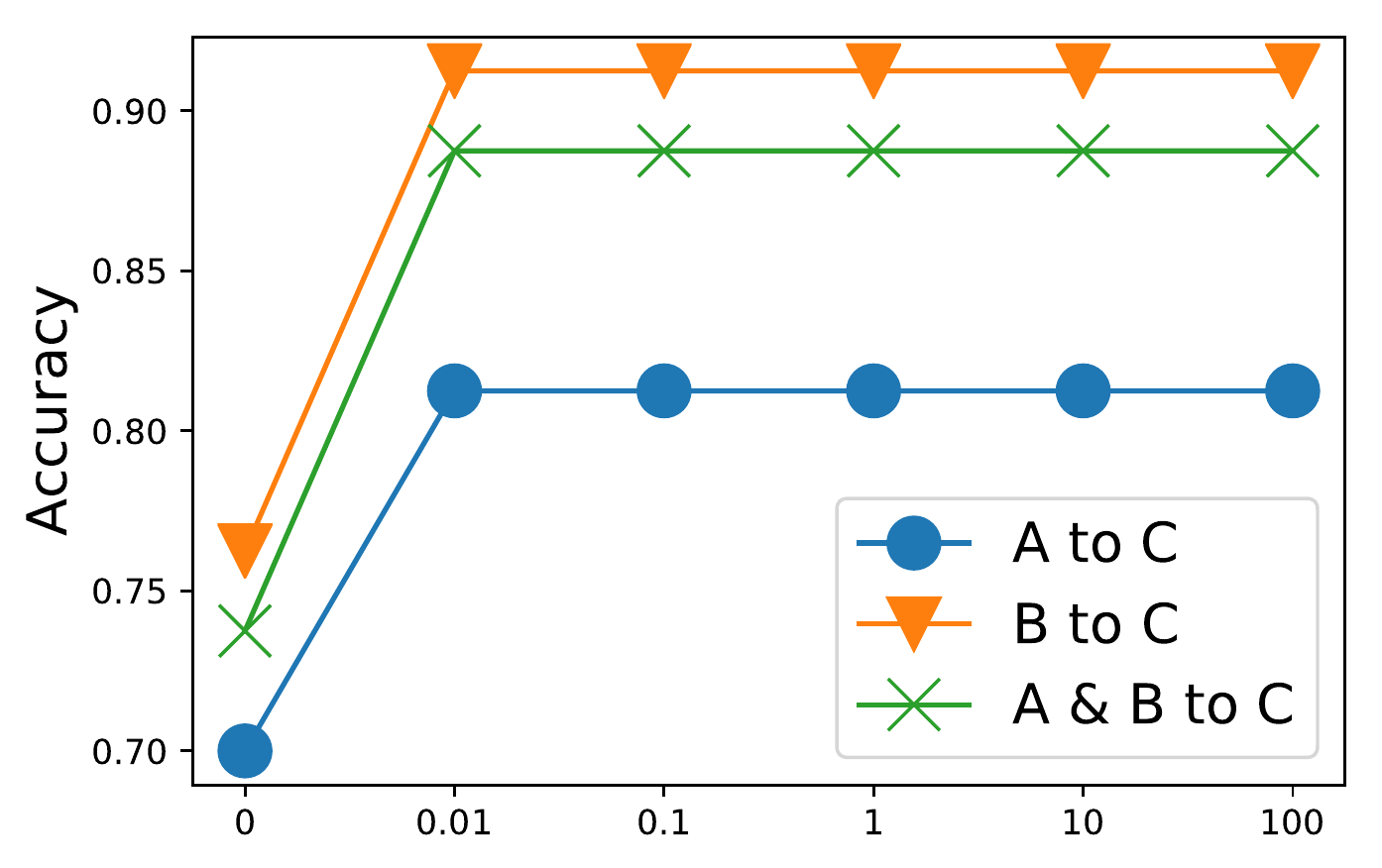}}
	\caption{The sensitivity of the classification accuracy with respect to hyper-parameters $C$ and $\lambda$ for DI-SVM with linear kernel.}
	\label{fig:param_sen}
\end{figure}
We evaluated the sensitivity of DI-SVM with linear kernel against hyper-parameters $ C $ and $ \lambda $ under five-fold cross validation.
Figure \ref{subfig:C} shows the sensitivity against $ C \in [10^{-3}, 10^4] $ when fixing $ \lambda = 1 $. We can observe that the accuracy stays stable when $ C \le 1 $, and shows a trend of decreasing when $ C \in [10^{0}, 10^4] $. Since a smaller value of $ C $ can lead to a larger SVM classification margin, we expect a classifier with a larger margin to generalize better and have higher prediction accuracy.
Figure \ref{subfig:lambda} shows the sensitivity against $ \lambda \in \{0, 0.01, 0.1, 1, 10, 100\} $ when fixing $ C = 1 $. We can observe that the prediction accuracy of DI-SVM kept almost constant when $ \lambda> 0 $ and it was not sensitive to $ \lambda $. When $ \lambda = 0 $, i.e., without minimizing domain dependence, DI-SVM becomes a standard Kernel SVM and the performance was poorer.

\section{Conclusion}
In this paper, we proposed Domain Independent SVM (DI-SVM) for transfer learning in brain decoding. It works by minimizing the SVM empirical risk and dependence on auxiliary domain information measured by a simplified HSIC. We evaluated DI-SVM against PCA and six state-of-the-art transfer learning methods on 13 transfer learning tasks. Experimental results showed the superior overall performance of DI-SVM over other methods, particularly on multi-source transfer with a 24\% improvement. This confirmed the benefits of leveraging auxiliary domain information and HSIC in transfer learning for brain condition classification.

\newpage
\bibliographystyle{named}
\bibliography{ijcai19}

\begin{thebibliography}{}

\bibitem[\protect\citeauthoryear{Aron \bgroup \em et al.\egroup
  }{2007}]{aron2007triangulating}
Adam~R Aron, Tim~E Behrens, Steve Smith, Michael~J Frank, and Russell~A
  Poldrack.
\newblock Triangulating a cognitive control network using diffusion-weighted
  magnetic resonance imaging (mri) and functional mri.
\newblock {\em Journal of Neuroscience}, 27(14):3743--3752, 2007.

\bibitem[\protect\citeauthoryear{Borgwardt \bgroup \em et al.\egroup
  }{2006}]{borgwardt2006integrating}
Karsten~M Borgwardt, Arthur Gretton, Malte~J Rasch, Hans-Peter Kriegel,
  Bernhard Sch{\"o}lkopf, and Alex~J Smola.
\newblock Integrating structured biological data by kernel maximum mean
  discrepancy.
\newblock {\em Bioinformatics}, 22(14):e49--e57, 2006.

\bibitem[\protect\citeauthoryear{Cao \bgroup \em et al.\egroup
  }{2018}]{cao2018unsupervised}
Yue Cao, Mingsheng Long, and Jianmin Wang.
\newblock Unsupervised domain adaptation with distribution matching machines.
\newblock In {\em Thirty-Second AAAI Conference on Artificial Intelligence},
  2018.

\bibitem[\protect\citeauthoryear{Chen \bgroup \em et al.\egroup
  }{2015}]{NIPS2015_5855}
Po-Hsuan~(Cameron) Chen, Janice Chen, Yaara Yeshurun, Uri Hasson, James Haxby,
  and Peter~J Ramadge.
\newblock A reduced-dimension {fMRI} shared response model.
\newblock In {\em Advances in Neural Information Processing Systems 28}, pages
  460--468. Curran Associates, Inc., 2015.

\bibitem[\protect\citeauthoryear{Chu \bgroup \em et al.\egroup
  }{2017}]{chu2017selective}
Wen-Sheng Chu, Fernando De~la Torre, and Jeffrey~F Cohn.
\newblock Selective transfer machine for personalized facial expression
  analysis.
\newblock {\em IEEE transactions on pattern analysis and machine intelligence},
  39(3):529--545, 2017.

\bibitem[\protect\citeauthoryear{Gorgolewski \bgroup \em et al.\egroup
  }{2017}]{gorgolewski2017openneuro}
Krzysztof Gorgolewski, Oscar Esteban, Gunnar Schaefer, Brian Wandell, and
  Russell Poldrack.
\newblock Openneuro: a free online platform for sharing and analysis of
  neuroimaging data.
\newblock In {\em Organization for Human Brain Mapping. Vancouver, Canada},
  page 1677, 2017.

\bibitem[\protect\citeauthoryear{Gretton \bgroup \em et al.\egroup
  }{2005}]{gretton2005measuring}
Arthur Gretton, Olivier Bousquet, Alex Smola, and Bernhard Sch{\"o}lkopf.
\newblock Measuring statistical dependence with hilbert-schmidt norms.
\newblock In {\em International conference on algorithmic learning theory},
  pages 63--77. Springer, 2005.

\bibitem[\protect\citeauthoryear{Kelly \bgroup \em et al.\egroup
  }{2008}]{kelly2008competition}
AM~Clare Kelly, Lucina~Q Uddin, Bharat~B Biswal, F~Xavier Castellanos, and
  Michael~P Milham.
\newblock Competition between functional brain networks mediates behavioral
  variability.
\newblock {\em Neuroimage}, 39(1):527--537, 2008.

\bibitem[\protect\citeauthoryear{Long \bgroup \em et al.\egroup
  }{2013}]{long2013jda}
Mingsheng Long, Jianmin Wang, Guiguang Ding, Jiaguang Sun, and Philip~S Yu.
\newblock Transfer feature learning with joint distribution adaptation.
\newblock In {\em Proceedings of the IEEE international conference on computer
  vision}, pages 2200--2207, 2013.

\bibitem[\protect\citeauthoryear{Mensch \bgroup \em et al.\egroup
  }{2017}]{mensch2017learning}
Arthur Mensch, Julien Mairal, Danilo Bzdok, Bertrand Thirion, and Ga{\"e}l
  Varoquaux.
\newblock Learning neural representations of human cognition across many fmri
  studies.
\newblock In {\em Advances in Neural Information Processing Systems}, pages
  5883--5893, 2017.

\bibitem[\protect\citeauthoryear{Ogawa \bgroup \em et al.\egroup
  }{1990}]{ogawa1990brain}
Seiji Ogawa, Tso-Ming Lee, Alan~R Kay, and David~W Tank.
\newblock Brain magnetic resonance imaging with contrast dependent on blood
  oxygenation.
\newblock {\em Proceedings of the National Academy of Sciences},
  87(24):9868--9872, 1990.

\bibitem[\protect\citeauthoryear{Pan and Yang}{2010}]{pan2010survey}
Sinno~Jialin Pan and Qiang Yang.
\newblock A survey on transfer learning.
\newblock {\em IEEE Transactions on knowledge and data engineering},
  22(10):1345--1359, 2010.

\bibitem[\protect\citeauthoryear{Pan \bgroup \em et al.\egroup
  }{2011}]{pan2011domain}
Sinno~Jialin Pan, Ivor~W Tsang, James~T Kwok, and Qiang Yang.
\newblock Domain adaptation via transfer component analysis.
\newblock {\em IEEE Transactions on Neural Networks}, 22(2):199--210, 2011.

\bibitem[\protect\citeauthoryear{Poldrack \bgroup \em et al.\egroup
  }{2013}]{poldrack2013toward}
Russell~A Poldrack, Deanna~M Barch, Jason Mitchell, Tor Wager, Anthony~D
  Wagner, Joseph~T Devlin, Chad Cumba, Oluwasanmi Koyejo, and Michael Milham.
\newblock Toward open sharing of task-based fmri data: the openfmri project.
\newblock {\em Frontiers in neuroinformatics}, 7:12, 2013.

\bibitem[\protect\citeauthoryear{Rao \bgroup \em et al.\egroup
  }{2013}]{rao2013sparse}
Nikhil Rao, Christopher Cox, Rob Nowak, and Timothy~T Rogers.
\newblock Sparse overlapping sets lasso for multitask learning and its
  application to fmri analysis.
\newblock In {\em Advances in neural information processing systems}, pages
  2202--2210, 2013.

\bibitem[\protect\citeauthoryear{Sch{\"o}lkopf \bgroup \em et al.\egroup
  }{2001}]{scholkopf2001generalized}
Bernhard Sch{\"o}lkopf, Ralf Herbrich, and Alex~J Smola.
\newblock A generalized representer theorem.
\newblock In {\em International conference on computational learning theory},
  pages 416--426. Springer, 2001.

\bibitem[\protect\citeauthoryear{Singh \bgroup \em et al.\egroup
  }{2007}]{Singh2007DetectionOC}
Vishwajeet Singh, Krishna~P. Miyapuram, and Raju~S. Bapi.
\newblock Detection of cognitive states from fmri data using machine learning
  techniques.
\newblock In {\em IJCAI 2007}, 2007.

\bibitem[\protect\citeauthoryear{Van~Essen \bgroup \em et al.\egroup
  }{2012}]{van2012human}
David~C Van~Essen, Kamil Ugurbil, E~Auerbach, D~Barch, TEJ Behrens, R~Bucholz,
  Acer Chang, Liyong Chen, Maurizio Corbetta, Sandra~W Curtiss, et~al.
\newblock The human connectome project: a data acquisition perspective.
\newblock {\em NeuroImage}, 62(4):2222--2231, 2012.

\bibitem[\protect\citeauthoryear{Wang \bgroup \em et al.\egroup
  }{2018}]{wang2018visual}
Jindong Wang, Wenjie Feng, Yiqiang Chen, Han Yu, Meiyu Huang, and Philip~S Yu.
\newblock Visual domain adaptation with manifold embedded distribution
  alignment.
\newblock In {\em 2018 ACM Multimedia Conference on Multimedia Conference},
  pages 402--410. ACM, 2018.

\bibitem[\protect\citeauthoryear{Weiss \bgroup \em et al.\egroup
  }{2016}]{weiss2016survey}
Karl Weiss, Taghi~M Khoshgoftaar, and DingDing Wang.
\newblock A survey of transfer learning.
\newblock {\em Journal of Big Data}, 3(1):9, 2016.

\bibitem[\protect\citeauthoryear{Xiao and Guo}{2015}]{xiao2015feature}
Min Xiao and Yuhong Guo.
\newblock Feature space independent semi-supervised domain adaptation via
  kernel matching.
\newblock {\em IEEE transactions on pattern analysis and machine intelligence},
  37(1):54--66, 2015.

\bibitem[\protect\citeauthoryear{Xue \bgroup \em et al.\egroup
  }{2008}]{xue2008common}
Gui Xue, Adam~R Aron, and Russell~A Poldrack.
\newblock Common neural substrates for inhibition of spoken and manual
  responses.
\newblock {\em Cerebral Cortex}, 18(8):1923--1932, 2008.

\bibitem[\protect\citeauthoryear{Yan \bgroup \em et al.\egroup
  }{2018}]{yan2018learning}
Ke~Yan, Lu~Kou, and David Zhang.
\newblock Learning domain-invariant subspace using domain features and
  independence maximization.
\newblock {\em IEEE transactions on cybernetics}, 48(1):288--299, 2018.

\bibitem[\protect\citeauthoryear{Zhang \bgroup \em et al.\egroup
  }{2018}]{zhang2018transfer}
Hejia Zhang, Po-Hsuan Chen, and Peter Ramadge.
\newblock Transfer learning on {fMRI} datasets.
\newblock In {\em Proceedings of Twenty-First International Conference on
  Artificial Intelligence and Statistics}, pages 595--603. PMLR, 2018.

\end{thebibliography}

\end{document}